\documentclass[sigconf]{acmart}

\AtBeginDocument{%
  \providecommand\BibTeX{{%
    \normalfont B\kern-0.5em{\scshape i\kern-0.25em b}\kern-0.8em\TeX}}}

\setcopyright{acmlicensed}
\copyrightyear{2024}
\acmYear{2024}
\acmDOI{}

\acmConference[Abbr]{Conf title}{MMM DD--DD,
  YYYY}{Location}
\acmISBN{978-1-4503-XXXX-X/18/06}

\usepackage{bm} 
\usepackage{amsfonts}
\usepackage{algorithm}
\usepackage{algpseudocode} 

\newcommand{\R}{\mathbb{R}}
\newcommand{\lrangle}[1]{\langle#1\rangle}
\newcommand{\lrvert}[1]{\lvert#1\rvert}
\newcommand{\lrVert}[1]{\lVert#1\rVert}
\newcommand{\lrVertF}[1]{\lVert#1\rVert_\text{F}}
\newcommand{\bG}{\bm\Gamma}
\newcommand{\bL}{\bm\Lambda}
\newcommand{\bY}{\bm Y}
\newcommand{\bX}{\bm X}
\newcommand{\bA}{\bm A}
\newcommand{\bB}{\bm B}

\newcommand{\bP}{\bm \Psi}
\newcommand{\bD}{\bm D}
\newcommand{\tR}{\tilde R}
\newcommand{\bbP}{\mathbb{P}}
\newcommand{\bT}{\bm T}

\newcommand{\bbPLp}{\mathbb{P}_{\mathcal{L}^\perp}}

\newcommand{\tdt}{\times\cdots\times}
\newcommand{\mat}[1]{_{[#1]}}
\newcommand{\optimes}[1]{\mathop{\times}#1}
\newcommand{\sumR}{\sum_{r_1=1}^{R_1}\cdots\sum_{r_q=1}^{R_q}}
\newcommand{\lr}{\lambda_{r_1,\ldots,r_q}}
\newcommand{\upp}[1]{^{(#1)}}
\newcommand{\vect}{\mathop{\mathrm{vec}}}

\newcommand{\diag}{\mathop{\mathrm{diag}}}

\begin{document}

\title{Oblivious subspace embeddings for compressed Tucker decompositions}

\author{Matthew Pietrosanu}
\affiliation{%
  \institution{Department of Mathematical \& Statistical Sciences, University of Alberta}
  \city{Edmonton}
  \country{Canada}}
\email{pietrosa@ualberta.ca}

\author{Bei Jiang}
\affiliation{%
  \institution{Department of Mathematical \& Statistical Sciences, University of Alberta}
  \city{Edmonton}
  \country{Canada}}
\email{bei1@ualberta.ca}

\author{Linglong Kong}
\affiliation{%
  \institution{Department of Mathematical \& Statistical Sciences, University of Alberta}
  \city{Edmonton}
  \country{Canada}}
\email{lkong@ualberta.ca}


\begin{abstract}
Emphasis in the tensor literature on random embeddings (tools for low-distortion dimension reduction) for the canonical polyadic (CP) tensor decomposition has left analogous results for the more expressive Tucker decomposition comparatively lacking. This work establishes general Johnson--Lindenstrauss (JL) type guarantees for the estimation of Tucker decompositions when an oblivious random embedding is applied along each mode. When these embeddings are drawn from a JL-optimal family, the decomposition can be estimated within $\varepsilon$ relative error under restrictions on the embedding dimension that are in line with recent CP results. We implement a higher-order orthogonal iteration (HOOI) decomposition algorithm with random embeddings to demonstrate the practical benefits of this approach and its potential to improve the accessibility of otherwise prohibitive tensor analyses. On moderately large face image and fMRI neuroimaging datasets, empirical results show that substantial dimension reduction is possible with minimal increase in reconstruction error relative to traditional\linebreak HOOI ($\leq$5\% larger error, 50\%--60\% lower computation time for large models with 50\% dimension reduction along each mode). Especially for large tensors, our method outperforms traditional higher-order singular value decomposition (HOSVD) and recently proposed TensorSketch methods.
\end{abstract}

\ccsdesc[500]{Mathematics of computing~Dimensionality reduction}
\ccsdesc[300]{Theory of computation~Sketching and sampling}

\keywords{alternating least squares, dimension reduction, higher-order orthogonal iteration, Johnson--Lindenstrauss, low-rank approximation, neuroimaging, random embedding, tensor decomposition}


\received{XX XXXXX 20XX}
\received[revised]{XX XXXXX 20XX}
\received[accepted]{XX XXXXX 20XX}

\maketitle

\section{Background and Motivation}


Low-dimensional decompositions lie at the heart of many tensor-based methods for statistical inference, representation, and feature extraction~\cite{tensor_txtbk}. Tensor-specialized methods are themselves often motivated by specific applied research questions and advancements in data-collection technologies (e.g., in pharmacology~\cite{pharm_tensor} and neuroimaging~\cite{neuro_tensor}). In these settings, naively working with vectorized data is typically neither conceptually sound nor computationally feasible. Dimension-reducing maps that preserve data geometry have consequently found substantial application in tensor research, where computational efficiency and representation quality are primary concerns due to prohibitive size of tensor data.

The well-known Johnson--Lindenstrauss (JL) lemma~\cite{jlt_proof} provides a theoretical basis for such maps. We say that a linear transformation $\bA$ is an $\varepsilon$-JL embedding of a set $\mathcal{S}\subset\R^n$ into $\R^m$ if, for every $x\in\mathcal{S}$, there exists $\varepsilon_x\in(-\varepsilon,\varepsilon)$ such that $$\lrVert{\bA x}_2^2 = (1+\varepsilon_x)\lrVert{x}_2^2.$$ JL-embeddings are typically generated randomly from a class of maps, independent of (or \emph{oblivious to}) the data to which it will be applied.

With much emphasis on scalable, computationally efficient tensor methods, it is not surprising that the majority of developments in the literature favor the canonical polyadic (CP) decomposition for its sparsity~\cite{iwen,trp,malik2020}. This has left theoretical results for the Tucker decomposition comparatively limited. For a given tensor $\bY\in\R^{n_1\tdt n_q}$, the Tucker decomposition takes the form 
\begin{align}
    \bY &= \sumR\lr\Gamma_{1,r_1}\circ\cdots\circ\Gamma_{q,r_q} \nonumber\\
    &=: [\bL\mid\bG_1,\ldots,\bG_q] \label{eq:tucker}
\end{align}
for a prespecified rank $(R_1,\ldots,R_q)$ ($R_j\leq n_j$), where $\bL=(\lr)$ $\in\R^{R_1\tdt R_q}$ is called the \emph{core tensor} and $\bG_j=[\Gamma_{j,1},\ldots,$ $\Gamma_{j,R_j}]\in\R^{n_j\times R_j}$ is called the $j$th \emph{factor matrix}. Briefly, the CP decomposition requires $\lr=0$ unless $r_1=\cdots=r_q$~\cite{kolda_review}. Though less sparse and computationally more complex, Tucker decompositions provide a richer class of decompositions and are the subject of continued research in statistics and other applied fields~\cite{tensor_txtbk,kolda_review,zhou_tucker}.

Recent notable work on Tucker decompositions by\linebreak Malik~\&~Becker~\cite{malik_nips}, Ma \& Solomonik~\cite{manips21}, and Minster~et~al.~\cite{minster} incorporate dimension reduction via CountSketch operators and random Gaussian matrices---both specific JL classes. The latter focuses on randomized higher-order SVD (HOSVD) for its computational benefits, whereas higher-order orthogonal iteration (HOOI) is known to provide better decompositions~\cite{hooi_vs_hosvd}. The work's randomized HOSVD and sequentially truncated HOSVD algorithms furthermore do not respect the Kronecker structure of the Tucker decomposition or only sequentially apply reductions along the data modes. Neither of these works provide general theoretical results that apply beyond a specific class of embeddings.

Similar comments apply to earlier literature that propose randomized Tucker decomposition algorithms (almost exclusively via HOSVD) through a specific class of JL embedding or take a different approach entirely (e.g., \cite{rand1,rand2,rand3}). For an overview and useful classification of randomized algorithms, see~\cite{rtucker_review}. These works highlight the wide interest in Tucker decompositions and the potential for application to a breath of problems, but again emphasize the lack of theoretical guarantees for general JL embedding frameworks.

This article considers the problem of estimating a Tucker decomposition of a given tensor $\bY\in\R^{n_1\tdt n_q}$ using a ``compressed'' version $\bY\times_1\bA_1\cdots\times_q\bA_q\in\R^{m_1\tdt m_q}$ of the data, where the $\bA_j$s are arbitrary JL embeddings. We focus solely on \emph{orthogonal Tucker decompositions}, namely, where the factor matrices have orthonormal columns (i.e., each $\bG_j$ lies on a Stiefel manifold). The two primary contributions of this work are as follows.
\begin{itemize}
    \item We establish JL-type theoretical results bounding the error introduced in exact and inexact Tucker decompositions when random embeddings from a JL-optimal family are applied along each data mode. We emphasize that these results apply generally to JL-optimal families and not to a specific type of embedding, unlike other works~\cite{malik_nips,minster}.
    \item We propose a new HOOI algorithm that uses random embeddings to estimate Tucker decompositions. Empirically, for large models, our approach requires substantially less computation time with only a small increase in reconstruction error relative to traditional HOOI and can make large tensor analyses feasible on moderate computing resources. Unlike other works~\cite{minster}, our approach takes advantage of the Kronecker structure of the Tucker decomposition and uses (nearly) fully compressed data in all updates of the estimated decomposition. Our approach outperforms HOSVD~\cite{kolda_review} and recently proposed TensorSketch methods~\cite{malik_nips} for Tucker decomposition.
\end{itemize}

Our approach closely follows part of recent substantial developments by Iwen et al.~\cite{iwen} for the CP decomposition, with an important distinction aside from the different decomposition. The authors' remark in Section 3.2 that CP results can be applied directly to Tucker decompositions is only true when the core tensor has a specific pattern of high sparsity (as the authors note, through an appropriate choice of tensor basis). A direct application of these previous results to general Tucker decompositions (due to the nested sums) violates an important basis incoherence requirement. Thus, the modified approach taken in this work is indeed necessary.

\section{Theoretical Results} 

\subsection{Notation}

As standard operations, let $\circ$ denote the tensor outer product, $\otimes$ the Kronecker product, and $\boxdot$ the Hadamard product. Let $\vect$ denote the usual vectorization operator. For a tensor $\bX\in\R^{n_1\tdt n_q}$, let $\bX\mat{j}\in\R^{n_j\times\prod_{k=1,k\neq j}^q n_j}$ be the mode-$j$ matricization of $\bX$. Vectorization and matricization are taken to be compatible in the sense that $\vect\bX = \vect\bX\mat{1}$.
Let $\times_j$ denote mode-$j$ multiplication of a tensor by a matrix. For notational convenience, we write $\bX\times_{j\in[q]}\bA_j$ to denote the repeated mode $j$ multiplication $\bX\times_1\bA_1\cdots\times_q\bA_q$ (where order of operations is irrelevant as multiplication along different modes commute).
For further detail on tensor operations, see Kolda \& Bader~\cite{kolda_review}. 
For tensors $\bX$ and $\bY$ of the same size, define the tensor inner product $\lrangle{\bX,\bY} = \lrangle{\vect \bX,\vect \bY}$ and the associated norm $\lrVert{\bX} = \lrVert{\vect\bX}_2$.

A notation for \emph{basis coherence}, introduced previously in Iwen et al.~\cite{iwen} for the CP decomposition (and arguably generalized here), will be convenient when studying the Tucker decomposition (but less useful conceptually due to the orthogonality restriction on the $\bG_j$s). Define the \emph{modewise coherence} of any decomposition of the form in Equation \ref{eq:tucker} as $\mu_{\bY} = \max_{j\in[q]}\mu_{\bY,j}$, where
$$\mu_{\bY,j} = \max_{\substack{k,h\in[R_j]\\k\neq h}}\frac{\lrvert{\lrangle{\Gamma_{j,k},\Gamma_{j,h}}}}{\lrVert{\Gamma_{j,k}}_2\lrVert{\Gamma_{j,h}}}_2$$
is called the \emph{mode-$j$ coherence}. As in other works~\cite{iwen}, owing to the nonuniqueness of CP/Tucker decompositions, we calculate coherence using the rank-$1$ terms of an explicitly given decomposition.

\subsection{Exact Decompositions Under JL Embeddings}

We begin our theoretical analysis with an elementary result on how Tucker decompositions are perturbed under arbitrary mode-$j$ multiplication. See the appendix for a proof of the following claim, which relies on routine manipulation and properties of tensor matricization.

\begin{lemma}\label{lem5}
    Let $j\in[q]$ and $\bB\in\R^{m_j\times n_j}$. Suppose that $\bY \in \R^{n_1\times\cdots\times n_q}$ has a rank-$(R_1,\ldots,R_q)$ Tucker decomposition $\bY = [\bm\Lambda\mid\bm\Gamma_1,\ldots,\bm\Gamma_q]$ and that $\min_{r\in [R_j]}\lrVert{\bB\Gamma_{j,r}}_2>0$. Then $$\bY^\prime = \bY\times_j\bB = [\bL^\prime\mid\bG_1^\prime,\ldots,\bG_q^\prime],$$ where $\lr^\prime=\lr\lrVert{\bB\Gamma_{j,r_j}}$, $\bG_k^\prime=\bG_k$ for $k\neq j$, and $\bG_j^\prime$ has columns $\Gamma_{j,r}/\lrVert{\bB\Gamma_{j,r}}$, $r\in[R_j]$.
    It follows that $\mu_{\bY^\prime,k}=\mu_{\bY,k}$ when $k\neq j$ and
    $$\mu_{\bY^\prime,j} = \max_{\substack{r,s\in[R_j]\\r\neq s}}\frac{\lrvert{\lrangle{\bB\Gamma_{j,r},\bB\Gamma_{j,s}}}}{\lrVert{\bB\Gamma_{j,r}}_2\lrVert{\bB\Gamma_{j,s}}_2}.$$ Furthermore,
    $$\lrVert{\bY^\prime}^2 = \sum_{r=1}^{R_j}\sum_{s=1}^{R_j}(\bP_j\bP_j^\top)_{r,s}\lrangle{\bB\Gamma_{j,r},\bB\Gamma_{j,s}},$$
    with $\bP_j = \bL\mat{j}\big(\bigotimes_{k=q,k\neq j}^1\bG_k\big)^\top\in\R^{R_j\times N_j}$ and $N_j = \prod_{k=1,k\neq j}^q n_k$.
\end{lemma}

When $\bB$ in Lemma \ref{lem5} is an $\varepsilon$-JL embedding that properly embeds the column space of the factor matrices of a Tucker decomposition, changes to the core tensor, coherence, and tensor norm can be controlled. This notion is formalized in Proposition \ref{prop1}.

\begin{proposition}\label{prop1}
    Fix $j\in[q]$ and suppose that $\bY\in\R^{n_1\tdt n_q}$ permits the rank-$(R_1,\ldots,R_q)$ Tucker decomposition $\bY = [\bL\mid\bG_1,\ldots,$ $\bG_q]$, where $\bG_j$ has columns of unit length. Suppose that $\bA\in\R^{m\times n_j}$ is an $\varepsilon$-JL embedding of the set
    $$\mathcal{S}_j = \bigcup_{\substack{r,s\in[R_j]\\r<s}}\{\Gamma_{j,r}\pm\Gamma_{j,s}\}\cup\bigcup_{r\in[R_j]}\{\Gamma_{j,r}\}\subset\R^{n_j}.$$
    Let $\bY^\prime = \bY\times_j\bA$, which (by Lemma \ref{lem5}) has the same decomposition as $\bY$ but with a core tensor $\bL^\prime$ with elements $\lr^\prime = \lr\lrVert{\bA\Gamma_{j,r_j}}_2$ and a $j$th factor matrix with columns $\bA\Gamma_{j,r_j}/$ $\lrVert{\bA\Gamma_{j,r_j}}_2$. Then
    \begin{enumerate}
        \item[(i)] $\lrvert{\lr^\prime-\lr}\leq\varepsilon\lrvert{\lr}$;
        \item[(ii)] $\mu_{\bY^\prime,j}\leq\varepsilon/(1-\varepsilon)$ and $\mu_{\bY^\prime,k}=\mu_{\bY,k}$ for all $k\neq j$; and
        \item[(iii)] $\big\lvert\lrVert{\bY^\prime}^2-\lrVert{\bY}^2\big\rvert \leq \big\lvert 1_{R_j}^\top[\bm E_j\boxdot(\bP_j\bP_j^\top)]1_{R_j}\big\rvert$,
    \end{enumerate}
    where $\bm E_j\in(-\varepsilon,\varepsilon)^{R_j\times R_j}$ and $\bP_j$ is as defined in Lemma \ref{lem5}.
\end{proposition}

\begin{proof}
    We provide a sketch of the proof. To prove claim (i), observe that
    \begin{align*}
        \lrvert{\lr^\prime-\lr} &= \lrvert{\lr}\big\lvert\lrVert{\bA\Gamma_{j,r_j}}_2-1\big\rvert\\
        &\leq \varepsilon\lrvert{\lr},
    \end{align*}
    where $\big\lvert\lrVert{\bA\Gamma_{j,r}}_2-1\big\rvert \leq \varepsilon$ since $\Gamma_{j,r}\in\mathcal{S}_j$ has unit norm.

    To prove claim (ii), note that, for any distinct $r$ and $s$,\linebreak $\lrvert{\lrangle{\bA\Gamma_{j,r},\bA\Gamma_{j,s}}-\lrangle{\Gamma_{j,r},\Gamma_{j,s}}}\leq\varepsilon$ by Lemma \ref{lem2} since $\{\Gamma_{j,r}\pm\Gamma_{j,s}\}\subset\mathcal{S}_j$. On the other hand,
    $$\lrVert{\bA\Gamma_{j,r}}_2\lrVert{\bA\Gamma_{j,s}}_2\geq\min_{t\in[R_j]}\lrVert{\bA\Gamma_{j,t}}_2^2\geq 1-\varepsilon$$
    as $\Gamma_{j,t}\in\mathcal{S}_j$. Thus,
    $$\mu_{\bY^\prime,j} = \max_{\substack{r,s\in[R_j]\\r\neq s}}\frac{\lrvert{\lrangle{\bA\Gamma_{j,r},\bA\Gamma_{j,s}}}}{\lrVert{\bA\Gamma_{j,r}}_2\lrVert{\bA\Gamma_{j,s}}_2}\leq \frac{\varepsilon}{1-\varepsilon}.$$
    It is clear that $\mu_{\bY^\prime,k}=\mu_{\bY,k}$ for $k\neq j$ since the $k$th factor matrix is the same between $\bY$ and $\bY^\prime$.

    Finally, to prove claim (iii), consider applications of Lemma \ref{lem5} with $\bB=\bA$ and $\bB=\bm I_{n_j}$. 
    Combining these representations of $\bY^\prime$ and $\bY$ yields claim (iii) and completes the proof of the proposition.
\end{proof}

Proposition \ref{prop1} controls the Tucker decomposition resulting from the application of a single embedding $\bA$ along mode $j$. Proposition \ref{prop2} repeatedly applies this result to obtain a JL-type bound for the application of an embedding $\bA_j$ along each mode.

\begin{proposition}\label{prop2}
    Let $\varepsilon\in(0,1)$ and suppose that $\bY \in \R^{n_1\times\cdots\times n_q}$ permits the rank-$(R_1,\ldots,R_q)$ orthogonal Tucker decomposition $\bY = [\bm\Lambda\mid\bm\Gamma_1,\ldots,\bm\Gamma_q]$. Let $\bA_j\in\R^{m_j\times n_j}$ be an $\varepsilon/q$-JL embedding of $\mathcal{S}_j$ (from Proposition \ref{prop1}) for each $j\in[q]$. Then
    $$\Big\lvert \lrVert{\bY}^2 - \big\lVert\bY\optimes{_{j=1}^q}\bA_j\big\rVert\Big\rvert \leq \frac{\tR\varepsilon e^{\varepsilon(2+\tR+2/q)}}{1+\varepsilon/(2q)}\lrVert{\bY}^2,$$
    where $\tR = \max_{j\in[q]}R_j$.
\end{proposition}

\begin{proof}
    We provide a sketch of the proof. For $j\in[q]$, define $\bY\upp{j} = \bY\upp{j-1}\times_j\bA_j = [\bL\upp{j}\mid\bG\upp{j}_1,\ldots,\bG\upp{j}_q]$, where $\lr\upp{j}=\lr\prod_{k=1}^j\lrVert{\bA_j\bG_{k,r_k}}_2$,
    \begin{align}
        \bG\upp{j}_k = \begin{cases}\bA_k\bG_k\bD_k^{-1}\quad&\text{if }k\leq j\\\bG_k&\text{if }k>j\end{cases},
    \end{align} 
    where $\bD_k=\diag(\{\lrVert{\bA_{k,r}\bG_{k,r}}_2\}_{r\in[R_k]})$ and $\bY\upp{0}=\bY$. Since $\bG_j\upp{j}$ has columns with unit norm, the results of Proposition \ref{prop1} can be applied to pairs of the form $(\bY\upp{j-1},\bY\upp{j})$.
    
    Toward the main result,
    \begin{align*}
        \Big\lvert \lrVert{\bY}^2 - \big\lVert\bY\optimes{_{j\in[q]}}\bA_j\big\rVert^2\Big\rvert
        &= \Big\lvert \sum_{j\in[q]} \big\lVert\bY\upp{j-1}\big\rVert^2 - \big\lVert\bY\upp{j}\big\rVert^2 \Big\rvert\\
        &\leq \sum_{j\in[q]} \big\lvert 1_{R_j}^\top[\bm E_j\boxdot(\bP_j\bP_j^\top)]1_{R_j}\big\rvert
    \end{align*}
    by claim (iii) of Proposition \ref{prop1}, here with $\bm E_j\in(-\varepsilon/q,\varepsilon/q)^{R_j\times R_j}$. Before proceeding, we must examine 
    the general term of the above sum:
    \begin{align*}
        \big\lvert &1_{R_j}^\top[\bm E_j\boxdot(\bP_j\bP_j^\top)]1_{R_j}\big\rvert \\
        &\leq \sum_{r=1}^{R_j}\sum_{s=1}^{R_j}\frac{\varepsilon}{q}\lrVert{e_r^\top\bL\mat{j}\upp{j}}_2\lrVert{e_s^\top\bL\mat{j}\upp{j}}_2\big\lVert\bigotimes_{\substack{k=q\\k\neq j}}^1\bG_k^{(j)\top}\bG_k\upp{j}\big\rVert_2.
    \end{align*}
    We calculate the final operator norm $\lrVert{\bigotimes_{k=q,k\neq j}^1\bG_k^{(j)\top}\bG_k\upp{j}}_2 = \prod_{k=q,k\neq j}^q\lrVert{\bG_k\upp{j}}_2^2$ in two cases. When $k>j$, it is clear that $\lrVert{\bG_k\upp{j}}_2^2 = 1$. On the other hand, when $k\leq j$,
    $$(\bG_k^{(j)\top}\bG\upp{j}_k)_{r,s} = \frac{\lrangle{\bA_k\Gamma_{k,r},\bA_k\Gamma_{k,s}}}{\lrVert{\bA_k\Gamma_{k,r}}_2\lrVert{\bA_k\Gamma_{k,s}}_2}$$
    for $r,s\in[R_k]$. This quantity is equal to $1$ when $r=s$ and, by claim (ii) of Proposition \ref{prop1}, is absolutely bounded by $\varepsilon/(q-\varepsilon)$ when $r\neq s$. 
    Thus, $\bG_k^{(j)\top}\bG\upp{j}_k = \bm I_{R_k} + (1_{R_k}1_{R_k}^\top - \bm I_{R_k})\boxdot\bm F_k$, where $\bm F$ has elements absolutely bounded by $\varepsilon/(q-\varepsilon)$.
    From this, it is straightforward to show that $\lrVert{\bG_k\upp{j}}_2^2 \leq 1 + \varepsilon(R_k-1)/(q-\varepsilon).$
    
    Returning to the main result, we can conclude that
    \begin{align*}
        &\Big\lvert \lrVert{\bY}^2 - \big\lVert\bY\optimes{_{j=1}^q}\bA_j\big\rVert^2\Big\rvert\\
        &\leq \frac{\varepsilon}{q}\sum_{j=1}^qR_j\Big(1+\frac{\varepsilon}{q}\Big)^{2j}\prod_{k=1}^{j-1}\Big[1+\Big(\frac{\varepsilon}{q-\varepsilon}\Big)(R_k-1)\Big]\lrVert{\bL}^2\\
        &\leq \frac{\tR\varepsilon e^{\varepsilon(2+\tR+2/q)}}{1+\varepsilon/(2q)}\lrVert{\bL}^2,
    \end{align*}
    where the final inequality holds since $\lrVert{\bL\upp{j}}^2 \leq (1+\varepsilon/q)^{2j}\lrVert{\bL}^2$ (by a recursive application of claim (i) of Proposition \ref{prop1}) and by other standard bounds. The result follows directly since $\lrVert{\bY} = \lrVert{\bL\times_{k\in[q]}\bG_k} = \lrVert{\bL}$.
\end{proof}

Before proceeding to the main result of this section, we introduce a convenient tool for the theoretical analysis of JL embeddings~\cite{iwen}. We say that a family of probability distributions $\mathcal{P}_{m,n}$ on $\R^{m\times n}$ over $(m,n)\in\mathbb{N}\times\mathbb{N}$ is an \emph{$\eta$-optimal family of JL embeddings} if there exists an absolute constant $C>0$ such that, for any $m<n$ and any set $\mathcal{S}\subset\R^n$ with cardinality $\lrvert{\mathcal{S}}\leq\eta\exp(\varepsilon^2m/C)$, the random matrix $\bA\sim\mathcal{P}_{m,n}$ is an $\varepsilon$-JL embedding of $\mathcal{S}$ into $\R^m$ with probability at least $1-\eta$. This concept permits general discussion of JL embeddings (as numerous optimal families exist).

\begin{theorem}\label{th1}
    Fix $\eta,\varepsilon\in(0,1)$ such that $\varepsilon\leq[\tR^{-1}+2^{-1}+(q\tR)^{-1}]^{-1}\ln2$. Let $\mathcal{L}=\mathrm{\mathrm{span}}\{\bigcirc_{j=1}^q\Gamma_{j,r_j}:r_j\in[R_j],j\in[q]\}$, where each $\bG_j\in\R^{n_j\times R_j}$ has orthonormal columns. Draw $\bA_j\in\R^{m_j\times n_j}$ from an $\eta/q$-optimal family of JL distributions, with
    $$m_j \geq \frac{\tilde C_j\tR^2q^2}{\varepsilon^2}\ln\Big(\frac{R_j^2q}{\eta}\Big)$$
    and where $\tilde C_j>0$ is some absolute constant. Then with probability at least $1-\eta$,
    $$\big\lvert\lrVert{\bY\optimes_{j=1}^q\bA_j}^2-\lrVert{\bY}^2\big\rvert \leq \varepsilon\lrVert{\bY}^2$$
    for all $\bY\in\mathcal{L}$.
\end{theorem}

\begin{proof}
    When drawn from an $\eta/q$-optimal family of JL distributions, $\bA_j$ is a $\delta/q$-JL embedding of $\mathcal{S}_j\subset\R^{n_j}$ into $\R^{m_j}$ with probability $1-\eta/d$ provided
    $$\lrvert{\mathcal{S}_j} = R_j^2 \leq \frac{\eta}{q}\exp\Big(\frac{(\delta/q)^2m_j}{C_j}\Big),$$
    where $C_j>0$ is some absolute constant. This sufficient condition is equivalent to $m_j\geq C_jq^2\delta^{-2}\ln(R_j^2q\eta^{-1})$.

    By Proposition \ref{prop2} (conditional on the $\bA_j$s being appropriate JL-embeddings),
    $$\Big\lvert \lrVert{\bY}^2 - \big\lVert\bY\optimes{_{j=1}^q}\bA_j\big\rVert\Big\rvert \leq \frac{\tR\delta e^{\delta(2+\tR+2/q)}}{1+\delta/(2q)}\lrVert{\bY}^2.$$
    Taking $\delta=\varepsilon/(2\tR)$ yields the desired sufficient condition on $m_j$ and allows the RHS above to be bounded by $\varepsilon\lrVert{\bY}^2$.
    The final part of the claim, the $1-\eta$ probability bound, holds by a union bound over $j\in[q]$.
\end{proof}

\subsection{Inexact Tucker Decompositions Under JL Embeddings}

Where the previous section concerned tensors with an exact Tucker decomposition, the following results consider approximate decompositions. The following lemms is a direct modification of Theorem 5 of \cite{iwen}, which we present here without proof.

\begin{lemma}\label{th5}
    Let $\bX\in\R^{n_1\tdt n_q}$ and $\bY\in\mathcal{L}\subset\mathop{\mathrm{span}}\{\bT_r\}_{r\in[R]}$, where the $\bT_r$s form an orthonormal set in $\R^{n_1\tdt n_q}$. Let $\bbPLp$ denote the orthogonal projection operator onto $\mathcal{L}^\perp$. Fix $\varepsilon\in(0,1)$ and let $L$ be a linear operator such that
    \begin{enumerate}
        \item[(i)] $L$ is an $\varepsilon/2$-JL embedding of $\mathcal{L}\cup\{\bbP_{\mathcal{L}^\perp}(\bX)\}$ into $\R^{m_1\tdt m_q}$, and
        \item[(ii)] $L$ is an $\varepsilon/(2\sqrt{R})$-JL embedding of $$\mathcal{S}^\prime = \bigcup_{r\in[R]}\Big\{\frac{\bbPLp(\bX)}{\lrVert{\bbPLp(\bX)}}\pm\bT_r\Big\}$$ into $\R^{m_1\tdt m_q}$.
    \end{enumerate}
    Then $\big\lvert \lrVert{L(\bX-\bY)}^2-\lrVert{\bX-\bY}^2\big\rvert \leq \varepsilon\lrVert{\bX-\bY}^2$ for all $\bY\in\mathcal{L}$.
\end{lemma}

Loosely, Lemma \ref{th5} provides conditions under which a linear operator can uniformly (over $\bY$ in some subspace of interest $\mathcal{L}$) embed $\bX-\bY$ for some arbitrary tensor $\bX$. This general result is needed for the major result in Theorem \ref{th4} below6a. .

\begin{theorem}\label{th4}
    For $q\geq2$, fix $\bX\in\R^{n_1\tdt n_q}$; $\eta,\varepsilon\in(0,1)$ with $\varepsilon\leq[(2\tR)^{-1}+4^{-1}+(2q\tR)^{-1}]^{-1}\ln 2$; and $j\in[q]$. Also fix $\bG_k\in\R^{n_k\times R_k}$, $k\in[q]\setminus\{j\}$, with orthonormal columns and $\bL\in\R^{R_1\tdt R_q}$. Define $\mathcal{L}_j = \{[\bL\mid\bG_1,\ldots,\bG_q]: \bG_j\in\R^{n_j\times R_j}, \bG_j^\top\bG_j= \bm I_{R_j}\}$. Let $\bA_k\in\R^{m_k\times n_k}$, $k\in[q]$, with
    $$m_k\geq \frac{\tilde C_jq^2\tilde p_j}{\varepsilon^2}\ln\Big(\frac{2D_j(q-1)}{\eta}\Big),$$
    be drawn from an $\eta/(2q)$-optimal family of JL distributions, where $\tilde C_j$ is an absolute constant, $D_j\geq(2\tilde p_j+1)\big(\prod_{\ell=1}^{k-1}m_\ell\big)\big(\prod_{\ell=k+1}^qn_\ell\big)$, and $\tilde p_j = \mathop{\mathrm{dim}}\mathop{\mathrm{span}}\mathcal{L}_j$. Define the random linear map $L$ via $L(\bm Z) = \bm Z\optimes_{k=1}^q\bA_k$. With probability at least $1-\eta$,
    $$\big\lvert \lrVert{L(\bX-\bY)}^2-\lrVert{\bX-\bY}^2\big\rvert \leq \varepsilon\lrVert{\bX-\bY}^2$$
    for all $\bY\in\mathcal{L}_j$. In particular, if $2R_j<n_j$, a sufficient condition for the embedding dimension is
    $$m_k\geq \frac{\tilde C_j(q+1)^3\tilde p_j}{\varepsilon^2}\ln\Big(\frac{4\tilde n}{\sqrt[\leftroot{1}\uproot{1}q+1]{\eta}}\Big),$$
    where $\tilde n = \max_{k\in[q]}n_k$.
\end{theorem}

\begin{proof}
    We present a sketch of the proof. Consider any $\bY\in\mathcal{L}_j$. By Proposition \ref{prop2} and an argument similar to that in Theorem \ref{th1}, drawing the $\bA_j$s from an $\varepsilon/(4\tR q)$-optimal JL family yields
    $$\big\lvert \lrVert{\bY}^2-\lrVert{L(\bY)}^2 \big\rvert \leq \frac{\varepsilon}{2}\lrVert{\bY}^2$$
    with probability $1-\eta/2$. The sufficient lower bound on $m_k$ is omitted as a tighter bound will be introduced shortly.
    
    
    Observe that
    $$\vect(\bY\mat{j}^\top) = \big\{\bm I_{n_j}\otimes\big[\big(\bigotimes_{\substack{k=q\\k\neq j}}^1\bG_k\big)\bL\mat{j}^\top\big]\big\}\bm C_j\gamma_j =: \bB_j\gamma_j$$
    where $C_j\gamma_j = \vect(\bG_j^\top)\in\R^{n_jR_j}$ represents $\vect(\bG_j^\top)\in\R^{n_jR_j}$ in terms of $\gamma_j\in\R^{p_j}$ (with $\bm C_j\in\R^{(n_jR_j)\times p_j}$ and $p_j < n_jR_j$), which is possible due to $\bG_j$'s orthogonal columns. Let $\tilde p_j\leq p_j$ denote the rank of $\bB_j$. It is clear that $\mathcal{L}_j$ is contained within a $\tilde p_j$-dimensional subspace of $\R^{n_1\tdt n_q}$. Let $\{\bT_r\}_{r\in[\tilde p_j]}$ be an orthonormal basis for this space.

    Now consider the $2\tilde p_j+1$ elements of
    $$\mathcal{S}_{j,0}^\prime = \{\bbPLp(\bX)\}\cup\bigcup_{r\in[\tilde p_j]}\Big\{\frac{\bbPLp(\bX)}{\lrVert{\bbPLp(\bX)}}\pm\bT_r\Big\}.$$
    Inductively define $\mathcal{S}_{j,k}^\prime = \{\bm Z\times_k\bA_k: \bm Z\in\mathcal{S}_{j,k-1}^\prime\}$. For each $k\in[q]$, $\bA_k$ is an $\varepsilon/(2eq\sqrt{\tilde p_j})$-JL embedding of the $(2\tilde p_j+1)\big(\prod_{\ell=1}^{k-1}m_\ell\big)\times$ $\big(\prod_{\ell=k+1}^qn_\ell\big)$ mode-$k$ fibers\footnote{A mode-$k$ fiber of a tensor $\bX\in\R^{n_1\tdt n_q}$ ($q\geq k$) is a vector of the form $(\bX_{n_1,\ldots,n_{k-1},i,n_{k+1},\ldots,n_q})_{i\in[n_k]}$, where $r_j\in[n_j]$ for $j\neq k$ are } of the elements of $\mathcal{S}_{j,k-1}^\prime$ with probability at least $1-\eta/(2q)$ provided that
    \begin{align*}
    (2\tilde p_j+1)&\big(\prod_{\ell=1}^{k-1}m_\ell\big)\big(\prod_{\ell=k+1}^qn_\ell\big)\\
    &\leq \frac{\eta}{2(q-1)}\exp\Big(\frac{[\varepsilon/(2eq\sqrt{\tilde p_j})]^2m_j}{C_j}\Big).
    \end{align*}
    Where $D_j$ is an upper bound for the LHS above, a sufficient condition for this result is
    $$m_k \geq \frac{\tilde C_jq^2\tilde p_j}{\varepsilon^2}\ln\Big(\frac{2D_j(q-1)}{\eta}\Big).$$
    
    By (a slight modification of) Lemma 9 of \cite{iwen}, it follows that $L$ is an $\varepsilon/(2\sqrt{\tilde p_j})$-JL embedding of $\mathcal{S}_j^\prime$ (with probability at least $1-\eta/2$).
    
    Lemma \ref{th5} thus applies and yields the desired result. The second bound follows under $2R_j<n_j$ by the bound $\sqrt[\leftroot{1}\uproot{1}q+1]{q}\leq\sqrt[\leftroot{1}\uproot{1}e]{e}\leq 2$.
\end{proof}

\section{Empirical Evaluation} \label{sec:empirical}

\subsection{Data and Setup}

We now present two real-world applications of random embeddings to the estimation of Tucker decompositions. Our main goal is to demonstrate empirically that compressed estimation can offer significant reductions in computation time in practice, even for moderately sized problems, with minimal impact on reconstruction error. All analyses were performed in MATLAB 2023a and use the implementation of modewise tensor multiplication in the Tensor Toolbox package (v3.5) on an Intel i7-8550U CPU with 16~GB of RAM.

The first analysis uses the ORL database of faces,\footnote{Available from AT\&T Laboratories Cambridge at \url{http://cam-orl.co.uk/facedatabase.html}} a collection of 400 grayscale images of size $92\times112$, featuring 40 subjects under 10 lighting conditions and with different facial expressions. We treat the dataset as a $92\times112\times400$ tensor and consider Tucker decompositions of rank $(R,R,R)$ for $R\in\{5,15,30\}$. The embedding dimension $(m_1,m_2,m_3)$ is controlled by a single (approximate) \emph{dimension reduction factor} $\mathrm{DR}$ (i.e., $\mathrm{DR}\approx m_j/n_j$ for all modes $j\in[q]$ to which compresion is applied).

The second analysis uses resting-state fMRI data, specifically for the hippocampus, obtained from the Alzheimer's Disease Neuroimaging Initiative (ADNI).\footnote{Data used in preparation of this article were obtained from the Alzheimer's Disease Neuroimaging Initiative (ADNI) database (adni.loni.usc.edu). As such, the investigators within the ADNI contributed to the design and implementation of ADNI and/or provided data but did not participate in analysis or writing of this report. A complete listing of ADNI investigators can be found at: \url{http://adni.loni.usc.edu/wp-content/uploads/how\_to\_apply/ADNI\_Acknowledgement\_List.pdf}} The data is a subset of that in Wang et al.~\cite{wang_data}, where details regarding data acquisition, extraction, and processing can be found. The hippocampal surface of each of the 824 subjects is parameterized as a $100\times150\times2$ tensor (with modes corresponding to rotation around the hippocampal surface, length along the surface, and the left/right hippocampus, respectively, as shown in Figure \ref{fig:param} in the appendix). Four surface-based fMRI measures, radial distance and three mTBM features, are available for each subject, for a $100\times150\times2\times4\times824$ data tensor. We consider Tucker decompositions of rank $(R, R, 2, 4, R)$ for $R\in\{5,15,30\}$ and reductions along the first, second, and last modes. 
No embeddings are applied to the third or fourth modes.

\begin{algorithm}[t]
\caption{Orthogonal Tucker decomposition via HOOI with random embeddings (HOOI-RE)}\label{alg:ALSRP}
\begin{algorithmic}
\Require{data $q$-tensor $\bX$; initial estimates $\bG,\ldots,\bG_q,\bL$; maximum iterations $N_\text{iter}$; relative tolerance $\varepsilon_\text{rel}$}
\State Initialize $\bm{D}_j$, $j\in[q]$
\State $\tilde\bX \gets \bX\times_{j\in[q]}(\bm{F}_j\bm{D}_j)$ \Comment{Mix}
\Repeat
    \State Generate $\bm{S}_j$, $j\in[q]$ \Comment{Form embedding}
    \For{$j\in[q]$} \Comment{Update $\bG_j$s}
        \State $\tilde\bX_j \gets \tilde\bX\times_{k\neq j}\bm{S}_k$ \Comment{Apply embedding}
        \State $\bG_j \gets \arg\min_{\bG_j}\lrVert{\tilde\bX_j-\bL\times_{k\neq j}(\bm{S}_k\bG_k)\times_j\bG_j}$
    \EndFor
    \State $\tilde\bX \gets \tilde\bX\times_{k\in[q]}\bm{S}_j$ \Comment{Apply embedding}
    \State $\bL \gets \arg\min_{\bL}\lrVert{\tilde\bX_j-\bL\times_{k\in[q]}(\bm{S}_k\bG_k)}$\\\Comment{Update $\bL$}
    \Until{relative fit does not improve by at least $\varepsilon_\text{rel}$} or until $N_\text{iter}$ iterations reached
\For{$j\in[q]$} \Comment{Unmix}
        \State $\bG_j \gets \bm{D}_j\bm{F}_j^\top\bG_j$
\EndFor
\State \Return $\bG_1,\ldots,\bG_q,\bL$
\end{algorithmic}
\end{algorithm}

\begin{figure*}[htb]
    \centering
    \includegraphics[width=\textwidth]{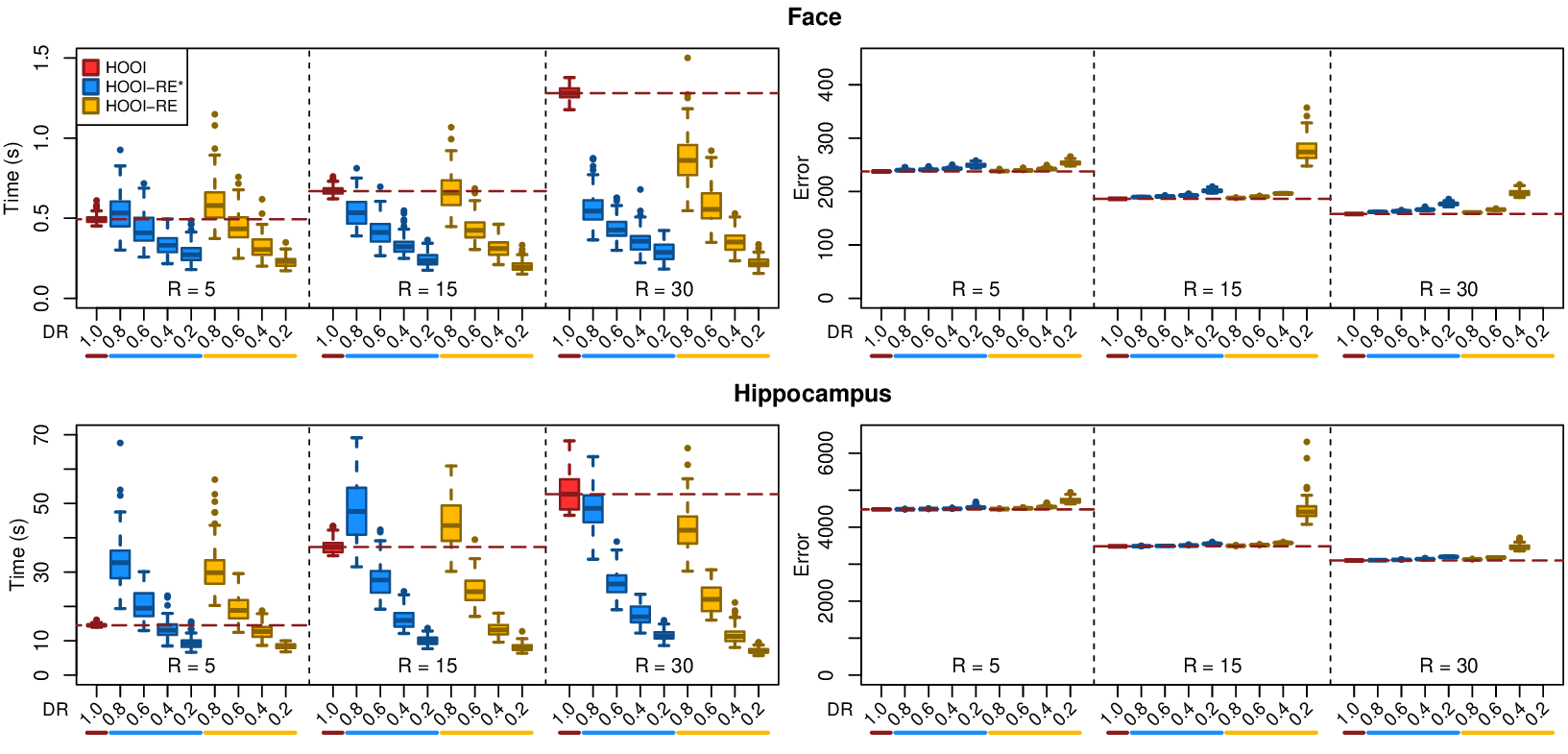}
    \caption{Total runtime and Frobenius reconstruction error for the HOOI, HOOI-RE*, and HOOI-RE algorithms in the face (left) and hippocampus (right) analyses, summarized over 100 replications. Horizontal dotted lines indicate median HOOI results. See Table \ref{tab:time_err} for HOOI-RE errors for $R=30$ and $\mathrm{DR}=0.2$, which are omitted here to avoid skewing the plot.}
    \Description{A four-panel figure. Each panel contains boxplots summarizing the total runtime or reconstruction error of the HOOI, HOOI-RE, and HOOI-RE* methods in the empirical analysis of the face or hippocampus data.}
    \label{fig:time_err}
\end{figure*}

In each analysis, we consider six estimation methods. The first is a traditional alternating least squares (ALS) algorithm, specifically, \emph{higher-order orthogonal iteration} (HOOI)~\cite{kolda_review}. The second, presented in Algorithm \ref{alg:ALSRP} as HOOI-RE,\footnote{Our implementation and code for the following numerical studies are available in an anonymized repository at \url{https://anonymous.4open.science/r/tucker_hooi-re-0CE3/README.md}} applies oblivious JL embeddings of the form $\bA_j = \bm{S}_j\bm{F}_j\bm{D}_j$ when updating each factor matrix $\bG_j$ and the core tensor $\bL$. Specifically, $\bm{S}_j\in\R^{m_j\times n_j}$ is a row-sampling matrix, $\bm{F}_j\in\R^{n_j\times n_j}$ is some orthogonal matrix (here a discrete cosine transformation matrix), and $\bm{D}_j\in\R^{n_j\times n_j}$ is a diagonal Rademacher matrix. The third algorithm, HOOI-RE*, is the same as HOOI-RE but uses the full data to estimate the core tensor: specifically, the core tensor update in HOOI-RE* seeks to minimize $\lrVert{\bX-\bL\times_{k\in[q]}\bG_k}$.

HOOI-RE applies $\bm{F}_j\bm{D}_j$ in an initial preprocessing step: loosely, this mixing step ``spreads'' information within the data $\bX$ and makes subsequent updates less sensitive to $\bm{S}_j$. Decomposing the mixed data $\tilde\bX$ is equivalent to decomposing $\bX$, so we need only ``unmix'' the estimates and return to the original $\bX$ space at the end. Algorithm \ref{alg:ALSRP} uses closed-form updates for tensor components, namely $\bG_j\gets\bm{U}_j\bm{V}_j^\top$, where $\bm{U}_j$ and $\bm{V}_j$ contain the left and right singular vectors from a thin SVD of $[\tilde\bX_j\times_{k\neq j}(\bm{S}_k\bG_k)]\mat{j}\bL\mat{j}^\top$, and $\bL \gets \tilde\bX\times_{j\in[q]}\{[(\bm{S}_j\bG_j)^\top(\bm{S}_j\bG_j)]^{-1}(\bm{S}_j\bG_j)^\top\}$ (where a pseudoinverse is used in the one case where $m_j<R_j$). Furthermore, the $\bm{S}_j$ matrices are formed and applied implicitly via subsetting rather than explicit matrix multiplication. HOOI, HOOI-RE, and HOOI-RE* use $\varepsilon_\text{rel}=1\times 10^{-5}$ and $N_\text{iter}=100$ (where the latter is never reached).

The last three methods use HOSVD \cite{kolda_review} and two TensorSketch algorithms (TUCKER-TS and TUCKER-TTMTS with default settings, as proposed and implemented in \cite{malik_nips}). They are presented for the sake of comparison to other (traditional and recent) approaches for estimating Tucker decompositions.

\subsection{Results}

Figure \ref{fig:time_err} visualizes the total computation time and final reconstruction error for the HOOI, HOOI-RE, and HOOI-RE* algorithms over 100 replications. Table \ref{tab:time_err} provides a numerical summary of the results for all six algorithms. For large decompositions, improvements in computation time are clear when $\mathrm{DR}<0.8$ for HOOI-RE and HOOI-RE*: $56\%$--$73\%$ reductions in the face analysis and $40\%$--$75\%$ reductions in the hippocampus analysis when $R=30$. At the same time, reconstruction error only suffers slightly: a $3\%$--$11\%$ increase in Frobenius reconstruction error in the same setting across both analyses.

Notable exceptions to this trend occur for HOOI-RE when $\mathrm{DR}=0.2$ or when $\mathrm{DR}=0.4$ and $R=30$---that is, where $m_j$ is small relative to $R_j$. In these settings, HOOI-RE yields a much higher reconstruction error than HOOI. A comparison of the results for HOOI-RE and HOOI-RE* suggest that the increased error is attributable to the instability of the least-squares update of $\bL$ when $m_j$ is close to or lower than $R_j$. In such extreme settings, HOOI-RE* retains good performance.

As shown in Table \ref{tab:time_err}, HOSVD and HOOI-RE are generally comparable in the face analysis. However, with the much larger hippocampus data, HOSVD takes substantially longer than HOOI-RE (even under minimal compression) and yields estimates with substantially larger reconstruction error. In both analyses, the two TensorSketch methods yield reconstruction errors higher than that for {HOOI-RE} with $\mathrm{DR}=0.4$. In most settings in the hippocampus analysis, the TensorSketch methods run out of memory (even when using variants of the algorithms that never hold the full data tensor in memory).

\begin{figure}[t]
    \centering
    \includegraphics[scale=0.8]{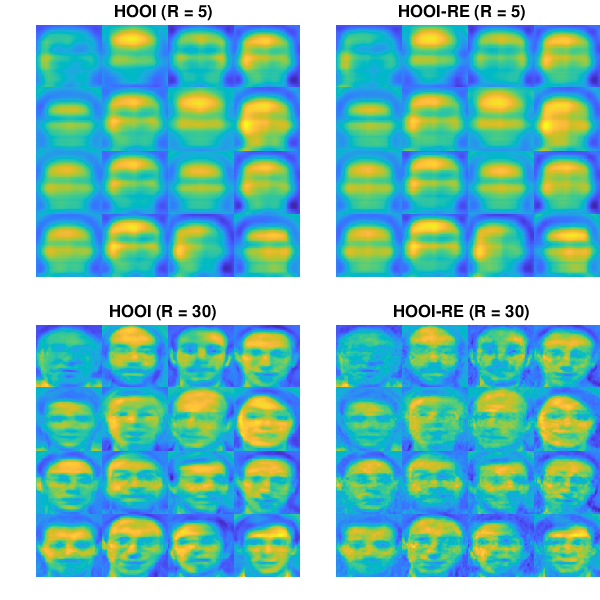}
    \caption{HOOI and HOOI-RE ($\mathrm{DR}=0.5$) reconstructions for 16 randomly selected faces in the face analysis. The original data is visualized in Figure \ref{fig:ALSRP}.}
    \Description{A four-panel figure. Each panel is divided into a four-by-four grid showing reconstructions of faces in the empirical analysis of the face dataset.}
    \label{fig:ALS_ALSRP}
\end{figure}

\begin{figure}[t]
    \centering
    \includegraphics[scale=0.8]{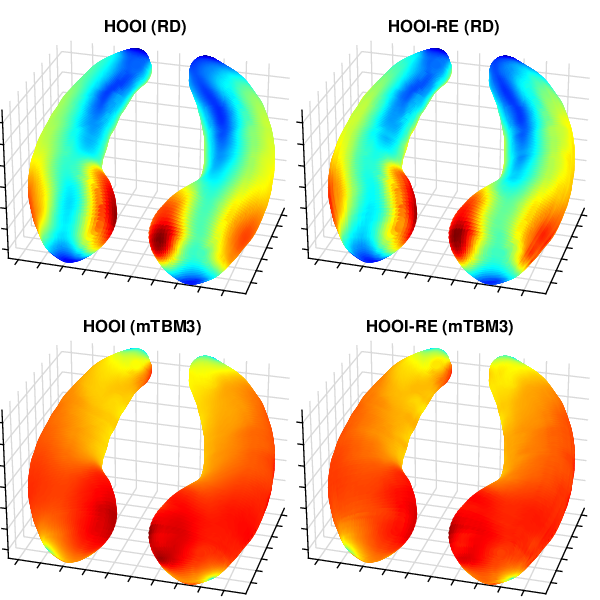}
    \caption{HOOI and HOOI-RE reconstructions ($R=30$, $\mathrm{DR}=0.5$) of the radial distance (RD) and mTBM3 imaging measures for a single patient in the hippocampus analysis.} 
    \Description{A four-panel figure. Each panel shows a 3D reconstruction of a hippocampus surface measure in the empirical analysis of the hippocampus dataset.}
    \label{fig:hippo}
\end{figure}

\begin{figure}[t]
    \centering
    \includegraphics[scale=0.75]{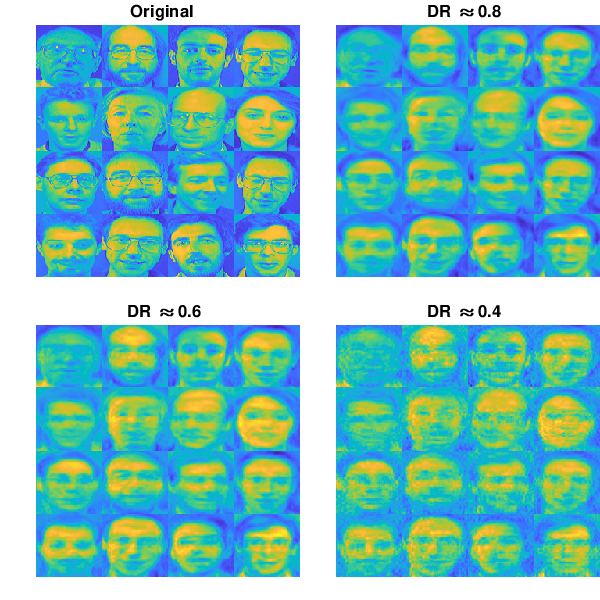}
    \caption{Effect of dimension reduction on HOOI-RE reconstructions in the face analysis ($R=30$) for 16 randomly selected images.}
    \label{fig:ALSRP}
	\Description{A four-panel figure. Each panel is divided into a four-by-four grid showing reconstructions of faces in the empirical analysis of the face dataset.}
\end{figure}

In the face analysis, for the same value of $R$ (which controls the coarseness of the decomposition), HOOI-RE appears to encode the same level of detail as HOOI, albeit with extra noise (Figure \ref{fig:ALS_ALSRP}). HOOI and HOOI-RE reconstructions in the hippocampus analysis are also comparable (Figure \ref{fig:hippo}). For an illustration of the effect of increased dimension reduction in the face analysis, see Figure \ref{fig:ALSRP}.

\begin{figure*}[t]
    \centering
    \includegraphics[width=\textwidth]{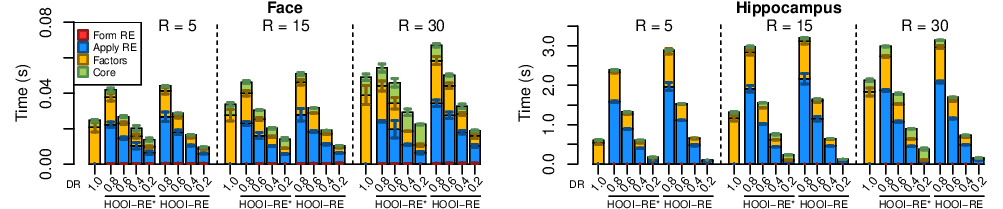}
    \caption{Average time spent per iteration on each part of the HOOI, HOOI-RE*, and HOOI-RE algorithms in the face (left) and hippocampus (right) analyses, averaged over 100 replications. Bars represent standard error. Preprocessing time is not included, but see Table \ref{tab:parts} in the appendix for quantitative summaries.}
    \Description{A two-panel figure. Each panel contains stacked barplots summarizing the average amount of time spent on each stage of the Tucker decomposition algorithm by the HOOI, HOOI-RE, and HOOI-RE* methods in the empirical analysis of the face or hippocampus data.}
    \label{fig:part}
\end{figure*}

For small models or when dimension reduction is not substantial (e.g., when $R=5$ or $\mathrm{DR}=0.8$), HOOI-RE and HOOI-RE* tend to require more computation time and, as before, yield greater reconstruction error than HOOI. The reason for this is clear from Figure \ref{fig:part}, which summarizes average time per iteration spent on each part of the HOOI algorithms (with numerical results in Table~\ref{tab:parts} of the appendix). Briefly, using embeddings incurs an overhead cost that may or may not be outweighed by the improvement in computation time needed to update the factor matrices or core tensor. There is a net improvement when the size of the decomposition is large or when the amount of dimension reduction is substantial (i.e., large $R$ or small $\mathrm{DR}$). HOSVD, like HOOI, spends a large majority of its runtime on updates to the factor matrices (Table~\ref{tab:parts} in the appendix). Our results highlight how the proposed approach with JL embeddings can reduce problem dimensionality and mitigate this computational bottleneck while preserving the integrity of the estimated decomposition components.

\section{Significance and Impact}

The importance of efficient methods for tensor analysis grows together with the size and richness of data available to researchers. This is especially true in applied fields where domain-specific insight is tethered to data acquisition and related technology. Medical imaging is a prime example of this. Data size may initially encourage researchers to reduce or altogether eliminate tensor data from an analysis plan---e.g., by summarizing neuroimaging features over predefined regions of interest or by downsampling to a manageable size, even when not statistically justifiable. However, tensor-based methods for dimension reduction can make large-scale analyses feasible on readily accessible computing resources. The empirical results in Section \ref{sec:empirical} show that our approach to HOOI can reduce problem size substantially and obtain high-quality Tucker decompositions in a fraction of the time. Relative to traditional HOOI and HOSVD, our method appears to scale well with data size $n$ and decomposition size $R$. The proposed method also substantially outperformed recently developed TensorSketch methods for HOOI in terms of reconstruction error and, particularly for larger tensors, computation time and computational feasibility.

While ``compressed'' tensor decompositions are not the only general tool needed, decompositions and low-rank approximations are arguably an important part of many tensor methods~\cite{tensor_txtbk}. Our results encourage further applications to tensor regression and other specialized methods, particularly those seeking a rich model/decomposition space through Tucker representations. There are settings where Tucker decompositions may be preferred over CP decompositions for reasons beyond flexibility. The latter requires the $R_j$s to be equal, but it may be more parsimonious to use a Tucker decomposition with greatly varying $R_j$s~\cite{zhou_tucker}. When analyzing our hippocampus data, for example, one may desire greater variability between patient-level reconstructions, so it may be preferable to have $R_5$ (along the ``patient'' mode) large and the other $R_j$s small. Expanded results for Tucker decompositions, such as those in this work, can thus support domain-specific developments even if corresponding results for CP decompositions exist.\footnote{While Theorem \ref{th1} appears to be restricted to $\bL$, Theorem \ref{th4} can be modified to account for this ($\bbPLp(\bX)=0$) case. Similarly, Theorem \ref{th4} can be easily modified to apply to $\bL$ by adjusting $\mathcal{L}_j$. We have omitted these details for brevity.}

\begin{table}[b!]
{\centering
\caption{Mean (standard deviation) runtime (s) and Frobenius reconstruction error for the HOOI, HOOI-RE, HOOI-RE*, HOSVD, and TensorSketch algorithms (with $K=10$) in both numerical studies, calculated over 100 simulations.} \label{tab:time_err}
\renewcommand{\arraystretch}{1} 
\setlength{\tabcolsep}{0.05em}
\scalebox{0.7}{
\begin{tabular}{llrrrrrr}
\hline
Method  & DR  & \multicolumn{3}{l}{Time}                    & \multicolumn{3}{l}{Error}                        \\ \cline{3-8} 
        &     & $R = 5$       & $R = 15$     & $R = 30$     & $R = 5$         & $R = 15$       & $R = 30$      \\ \midrule
\multicolumn{8}{c}{Face}                                                                                       \\ \hline
HOOI     &  & 0.47 (0.02)   & 0.57 (0.01)  & 1.13 (0.04)  & 237.5 (0.0)     & 186.4 (0.0)    & 158.1 (0.0)   \\
HOOI-RE* & 0.8 & 0.58 (0.14)   & 0.55 (0.10)  & 0.60 (0.12)  & 240.0 (0.8)     & 189.6 (0.5)    & 161.9 (0.5)   \\
        & 0.6 & 0.44 (0.08)   & 0.42 (0.07)  & 0.47 (0.07)  & 241.2 (1.8)     & 190.6 (0.6)    & 163.2 (0.6)   \\
        & 0.4 & 0.39 (0.07)   & 0.34 (0.06)  & 0.38 (0.07)  & 242.9 (1.7)     & 193.0 (1.0)    & 166.1 (1.1)   \\
        & 0.2 & 0.32 (0.06)   & 0.27 (0.04)  & 0.31 (0.06)  & 249.4 (3.1)     & 201.5 (2.3)    & 176.3 (2.6)   \\
HOOI-RE  & 0.8 & 0.68 (0.16)   & 0.70 (0.13)  & 0.78 (0.15)  & 238.3 (0.5)     & 188.0 (0.2)    & 160.8 (0.2)   \\
        & 0.6 & 0.49 (0.12)   & 0.48 (0.08)  & 0.50 (0.09)  & 239.7 (1.2)     & 190.4 (0.4)    & 166.3 (0.5)   \\
        & 0.4 & 0.38 (0.09)  & 0.34 (0.06)  & 0.30 (0.06)  & 242.4 (1.7)     & 196.5 (0.8)    & 197.3 (5.1)   \\
        & 0.2 & 0.28 (0.05) & 0.24 (0.04)  & 0.19 (0.04)  & 252.5 (3.2)     & 284.7 (32.9)   & 831.9 (42.5)  \\
HOSVD   &     & 0.45 (0.04)   & 0.45 (0.04)  & 0.45 (0.05)  & 241.0 (0.0)     & 186.8 (0.0)    & 158.2 (0.0)   \\
TS & & 0.25 (0.01) & 106.24 (9.42) &${}^\dag$ & 262.9 (4.0) & 199.7 (0.4) &${}^\dag$\\
TTMTS & & 0.19 (0.01) & 2.97 (0.06) & 144.22 (3.24) & 480.7 (26.4) & 394.1 (6.9) & 370.7 (2.8)\\
\hline
\multicolumn{8}{c}{Hippocampus}                                                                                \\ \hline
HOOI     &  & 14.60 (0.28)  & 34.01 (0.28) & 44.74 (0.41) & 4480.1 (0.0)    & 3481.5 (0.0)   & 3099.2 (0.0)  \\
HOOI-RE* & 0.8 & 33.90 (7.41)  & 42.72 (8.05) & 46.37 (6.62) & 4483.1 (1.8)    & 3486.3 (1.2)   & 3106.2 (1.4)  \\
        & 0.6 & 20.88 (3.75)  & 24.81 (4.02) & 26.69 (3.57) & 4487.7 (2.4)    & 3492.7 (2.7)   & 3116.2 (2.5)  \\
        & 0.4 & 13.07 (2.17)  & 15.03 (2.27) & 16.43 (2.36) & 4499.5 (9.8)    & 3504.8 (4.6)   & 3135.4 (4.7)  \\
        & 0.2 & 9.23 (1.20)   & 9.32 (1.19)  & 10.85 (1.48) & 4533.8 (22.8) & 3546.3 (11.2)  & 3200.5 (12.9) \\
HOOI-RE  & 0.8 & 29.98 (7.11)  & 40.31 (5.96) & 44.47 (7.56) & 4490.6 (2.9)    & 3494.3 (2.1)   & 3124.7 (2.6)  \\
        & 0.6 & 18.14 (3.69)  & 21.30 (3.18) & 22.35 (3.71) & 4508.4 (10.5)   & 3516.1 (3.5)   & 3179.9 (9.5)  \\
        & 0.4 & 11.68 (2.02)  & 12.26 (1.91) & 11.16 (1.68) & 4550.7 (25.1)   & 3574.7 (11.1)  & 3466.8 (58.8) \\
        & 0.2 & 7.28 (0.57)  & 7.28 (0.89)  & 6.68 (0.72)  & 4712.4 (69.1)   & 4462.7 (292.7) & 9900.4 (11.2) \\ 
HOSVD   &     & 42.35 (1.03) & 42.35 (0.87) & 42.94 (1.18) & 5856.1 (0.0) & 4446.4 (0.0) & 3936.4 (0.0) \\
TS    & & 42.52 (1.78)${}^\dag$ & ${}^\dag$ & ${}^\dag$ & 4790.0 (12.5) & ${}^\dag$ & ${}^\dag$ \\
TTMTS & & 6.63 (0.11) & ${}^\dag$ &  ${}^\dag$ & 6421.4 (87.5)& ${}^\dag$ & ${}^\dag$\\
\hline
\end{tabular}
}}
{\\\footnotesize $\dag$ indicates that the TS or TTMTS algorithm ran out of memory. A version of the algorithm that does not hold data in memory~\cite{malik_nips} was used instead where possible, but in most cases also ran out of memory.}
\end{table}

We considered an HOOI algorithm in this work partly because of its rarity in the literature relative to HOSVD. More specifically, while HOSVD is typically favored for computational speed~\cite{hooi_vs_hosvd}, HOOI (and thus, algorithms that widen the applicability of HOOI to large tensor data) are of specific interest for the sake for improved decomposition quality. Our results in fact show that the proposed HOOI method can outperform HOSVD in terms of both computation time and decomposition quality, with the performance gap growing with the size and order of the data. The majority of the improvement in computation time per iteration appears to stem from improvements in factor matrix updates, in turn due to faster SVDs. Randomized SVD has itself received much attention in the broader tensor literature~\cite{rtucker_review}.

We acknowledge that further improvement in runtime is possible by fixing the sampling matrices (i.e., the $\bm{S}_j$s) across iterations (similar to the fixed sketches in \cite{malik_nips}), but this in our experience tends to increase reconstruction error and variation in reconstruction quality. We did not investigate this theoretically or present related empirical results in this work. Our choice to vary the $\bm{S}_j$s was primarily motivated by our interest in this source of randomness, from the perspective of posterior approxmation in Bayesian settings~\cite{pmex}. Specifically, work on Bayesian hierarchical tensor modeling is currently quite limited: while very few works consider dimension reduction for Bayesian regression with non-tensor data~\cite{bayes_comp_reg,guhaniyogi2021sketching}, none consider this in the context of joint tensor models. We are currently developing a compressed Bayesian Tucker decomposition that we ultimately aim to incorporate into joint hierarchical models.

We did not consider a second stage of dimension reduction (e.g., by vectorizing and embedding the compressed tensor into a low-dimensional vector space such as $\R^m$) as in Iwen et al.~\cite{iwen} for the CP decomposition. A similar result for the Tucker decomposition may follow readily, but we have not considered that here and leave the development for future work. In another vein, theoretical convergence guarantees are generally difficult to obtain for ALS algorithms (including HOOI, even without random embeddings)~\cite{kolda_review}, so we have not considered such results here, nor have we considered a formal runtime analysis.

\bibliographystyle{ACM-Reference-Format}
\bibliography{sample-authordraft}


\begin{thebibliography}{21}


\ifx \showCODEN    \undefined \def \showCODEN     #1{\unskip}     \fi
\ifx \showDOI      \undefined \def \showDOI       #1{#1}\fi
\ifx \showISBNx    \undefined \def \showISBNx     #1{\unskip}     \fi
\ifx \showISBNxiii \undefined \def \showISBNxiii  #1{\unskip}     \fi
\ifx \showISSN     \undefined \def \showISSN      #1{\unskip}     \fi
\ifx \showLCCN     \undefined \def \showLCCN      #1{\unskip}     \fi
\ifx \shownote     \undefined \def \shownote      #1{#1}          \fi
\ifx \showarticletitle \undefined \def \showarticletitle #1{#1}   \fi
\ifx \showURL      \undefined \def \showURL       {\relax}        \fi
\providecommand\bibfield[2]{#2}
\providecommand\bibinfo[2]{#2}
\providecommand\natexlab[1]{#1}
\providecommand\showeprint[2][]{arXiv:#2}

\bibitem[Ahmadi-Asl et~al\mbox{.}(2021)]%
        {rtucker_review}
\bibfield{author}{\bibinfo{person}{Salman Ahmadi-Asl},
  \bibinfo{person}{Stanislav Abukhovich}, \bibinfo{person}{Maame~G.
  Asante-Mensah}, \bibinfo{person}{Andrzej Cichocki}, \bibinfo{person}{Anh~Huy
  Phan}, \bibinfo{person}{Tohishisa Tanaka}, {and} \bibinfo{person}{Ivan
  Oseledets}.} \bibinfo{year}{2021}\natexlab{}.
\newblock \showarticletitle{Randomized algorithms for computation of {T}ucker
  decomposition and higher order {SVD} ({HOSVD})}.
\newblock \bibinfo{journal}{\emph{IEEE Access}}  \bibinfo{volume}{9}
  (\bibinfo{year}{2021}), \bibinfo{pages}{28684--28706}.
\newblock
\urldef\tempurl%
\url{https://doi.org/10.1109/ACCESS.2021.3058103}
\showDOI{\tempurl}


\bibitem[da~Costa et~al\mbox{.}(2016)]%
        {rand2}
\bibfield{author}{\bibinfo{person}{Michele~N. da Costa},
  \bibinfo{person}{Renato~R. Lopes}, {and} \bibinfo{person}{Jo\=ao Marcos~T.
  Romano}.} \bibinfo{year}{2016}\natexlab{}.
\newblock \showarticletitle{Randomized methods for higher-order subspace
  separation}. In \bibinfo{booktitle}{\emph{2016 24th European Signal
  Processing Conference (EUSIPCO)}}. \bibinfo{pages}{215--219}.
\newblock
\urldef\tempurl%
\url{https://doi.org/10.1109/EUSIPCO.2016.7760241}
\showDOI{\tempurl}


\bibitem[Dasgupta and Gupta(2003)]%
        {jlt_proof}
\bibfield{author}{\bibinfo{person}{Sanjoy Dasgupta} {and}
  \bibinfo{person}{Anupam Gupta}.} \bibinfo{year}{2003}\natexlab{}.
\newblock \showarticletitle{An elementary proof of a theorem of {J}ohnson and
  {L}indenstrauss}.
\newblock \bibinfo{journal}{\emph{Random Structures \& Algorithms}}
  \bibinfo{volume}{22}, \bibinfo{number}{1} (\bibinfo{year}{2003}),
  \bibinfo{pages}{60--65}.
\newblock
\urldef\tempurl%
\url{https://doi.org/10.1002/rsa.10073}
\showDOI{\tempurl}
\showeprint{https://onlinelibrary.wiley.com/doi/pdf/10.1002/rsa.10073}


\bibitem[De~Lathauwer et~al\mbox{.}(2000)]%
        {hooi_vs_hosvd}
\bibfield{author}{\bibinfo{person}{Lieven De~Lathauwer}, \bibinfo{person}{Bart
  De~Moor}, {and} \bibinfo{person}{Joos Vandewalle}.}
  \bibinfo{year}{2000}\natexlab{}.
\newblock \showarticletitle{On the best rank-1 and rank-$({R}_1,
  {R}_2,\ldots,{R}_N)$ approximation of higher-order tensors}.
\newblock \bibinfo{journal}{\emph{SIAM J. Matrix Anal. Appl.}}
  \bibinfo{volume}{21}, \bibinfo{number}{4} (\bibinfo{year}{2000}),
  \bibinfo{pages}{1324--1342}.
\newblock
\urldef\tempurl%
\url{https://doi.org/10.1137/S0895479898346995}
\showDOI{\tempurl}


\bibitem[Drineas and Mahoney(2007)]%
        {rand1}
\bibfield{author}{\bibinfo{person}{Petros Drineas} {and}
  \bibinfo{person}{Michael~W. Mahoney}.} \bibinfo{year}{2007}\natexlab{}.
\newblock \showarticletitle{A randomized algorithm for a tensor-based
  generalization of the singular value decomposition}.
\newblock \bibinfo{journal}{\emph{Linear Algebra Appl.}} \bibinfo{volume}{420},
  \bibinfo{number}{2} (\bibinfo{year}{2007}), \bibinfo{pages}{553--571}.
\newblock
\showISSN{0024--3795}
\urldef\tempurl%
\url{https://doi.org/10.1016/j.laa.2006.08.023}
\showDOI{\tempurl}


\bibitem[Gachloo et~al\mbox{.}(2019)]%
        {pharm_tensor}
\bibfield{author}{\bibinfo{person}{Mina Gachloo}, \bibinfo{person}{Yuxing
  Wang}, {and} \bibinfo{person}{Jingbo Xia}.} \bibinfo{year}{2019}\natexlab{}.
\newblock \showarticletitle{A review of drug knowledge discovery using
  {B}io{NLP} and tensor or matrix decomposition}.
\newblock \bibinfo{journal}{\emph{Genomics \& Informatics}}
  \bibinfo{volume}{17}, \bibinfo{number}{2} (\bibinfo{year}{2019}),
  \bibinfo{pages}{e18}.
\newblock


\bibitem[Guhaniyogi and Dunson(2015)]%
        {bayes_comp_reg}
\bibfield{author}{\bibinfo{person}{Rajarshi Guhaniyogi} {and}
  \bibinfo{person}{David~B. Dunson}.} \bibinfo{year}{2015}\natexlab{}.
\newblock \showarticletitle{Bayesian compressed regression}.
\newblock \bibinfo{journal}{\emph{J. Amer. Statist. Assoc.}}
  \bibinfo{volume}{110}, \bibinfo{number}{512} (\bibinfo{year}{2015}),
  \bibinfo{pages}{1500--1514}.
\newblock
\urldef\tempurl%
\url{https://doi.org/10.1080/01621459.2014.969425}
\showDOI{\tempurl}


\bibitem[Guhaniyogi and Scheffler(2021)]%
        {guhaniyogi2021sketching}
\bibfield{author}{\bibinfo{person}{Rajarshi Guhaniyogi} {and}
  \bibinfo{person}{Aaron Scheffler}.} \bibinfo{year}{2021}\natexlab{}.
\newblock \bibinfo{title}{Sketching in {B}ayesian high dimensional regression
  with big data using {G}aussian scale mixture priors}.
\newblock
\newblock
\showeprint[arxiv]{2105.04795}~[stat.ME]


\bibitem[Iwen et~al\mbox{.}(2021)]%
        {iwen}
\bibfield{author}{\bibinfo{person}{Mark~A. Iwen}, \bibinfo{person}{Deanna
  Needell}, \bibinfo{person}{Elizaveta Rebrova}, {and} \bibinfo{person}{Ali
  Zare}.} \bibinfo{year}{2021}\natexlab{}.
\newblock \showarticletitle{Lower memory oblivious (tensor) subspace embeddings
  with fewer random bits: Modewise methods for least squares}.
\newblock \bibinfo{journal}{\emph{SIAM J. Matrix Anal. Appl.}}
  \bibinfo{volume}{42}, \bibinfo{number}{1} (\bibinfo{year}{2021}),
  \bibinfo{pages}{376--416}.
\newblock
\urldef\tempurl%
\url{https://doi.org/10.1137/19M1308116}
\showDOI{\tempurl}


\bibitem[Kolda and Bader(2009)]%
        {kolda_review}
\bibfield{author}{\bibinfo{person}{Tamara~G. Kolda} {and}
  \bibinfo{person}{Brett~W. Bader}.} \bibinfo{year}{2009}\natexlab{}.
\newblock \showarticletitle{Tensor decompositions and applications}.
\newblock \bibinfo{journal}{\emph{SIAM Rev.}} \bibinfo{volume}{51},
  \bibinfo{number}{3} (\bibinfo{year}{2009}), \bibinfo{pages}{455--500}.
\newblock
\urldef\tempurl%
\url{https://doi.org/10.1137/07070111X}
\showDOI{\tempurl}


\bibitem[Li et~al\mbox{.}(2018)]%
        {zhou_tucker}
\bibfield{author}{\bibinfo{person}{Xiaoshan Li}, \bibinfo{person}{Da Xu},
  \bibinfo{person}{Hua Zhou}, {and} \bibinfo{person}{Lexin Li}.}
  \bibinfo{year}{2018}\natexlab{}.
\newblock \showarticletitle{Tucker tensor regression and neuroimaging
  analysis}.
\newblock \bibinfo{journal}{\emph{Statistics in Biosciences}}
  \bibinfo{volume}{10}, \bibinfo{number}{3} (\bibinfo{year}{2018}),
  \bibinfo{pages}{520--545}.
\newblock
\urldef\tempurl%
\url{https://doi.org/10.1007/s12561-018-9215-6}
\showDOI{\tempurl}


\bibitem[Liu and Wu(1999)]%
        {pmex}
\bibfield{author}{\bibinfo{person}{Jun~S. Liu} {and} \bibinfo{person}{Ying~Nian
  Wu}.} \bibinfo{year}{1999}\natexlab{}.
\newblock \showarticletitle{Parameter expansion for data augmentation}.
\newblock \bibinfo{journal}{\emph{J. Amer. Statist. Assoc.}}
  \bibinfo{volume}{94}, \bibinfo{number}{448} (\bibinfo{year}{1999}),
  \bibinfo{pages}{1264--1274}.
\newblock


\bibitem[Liu(2022)]%
        {tensor_txtbk}
\bibfield{author}{\bibinfo{person}{Yipeng Liu}.}
  \bibinfo{year}{2022}\natexlab{}.
\newblock \bibinfo{booktitle}{\emph{Tensors for Data Processing: Theory,
  Methods and Applications}}.
\newblock \bibinfo{publisher}{Elsevier}.
\newblock


\bibitem[Ma and Solomonik(2021)]%
        {manips21}
\bibfield{author}{\bibinfo{person}{Linjian Ma} {and} \bibinfo{person}{Edgar
  Solomonik}.} \bibinfo{year}{2021}\natexlab{}.
\newblock \showarticletitle{Fast and accurate randomized algorithms for
  low-rank tensor decompositions}. In \bibinfo{booktitle}{\emph{Advances in
  Neural Information Processing Systems}},
  \bibfield{editor}{\bibinfo{person}{A.~Beygelzimer},
  \bibinfo{person}{Y.~Dauphin}, \bibinfo{person}{P.~Liang}, {and}
  \bibinfo{person}{J.~Wortman Vaughan}} (Eds.).
\newblock
\urldef\tempurl%
\url{https://openreview.net/forum?id=B4szfz7W7LU}
\showURL{%
\tempurl}


\bibitem[Malik and Becker(2018)]%
        {malik_nips}
\bibfield{author}{\bibinfo{person}{Osman~Asif Malik} {and}
  \bibinfo{person}{Stephen Becker}.} \bibinfo{year}{2018}\natexlab{}.
\newblock \showarticletitle{Low-rank {T}ucker decomposition of large tensors
  using {T}ensor{S}ketch}. In \bibinfo{booktitle}{\emph{Advances in Neural
  Information Processing Systems}},
  \bibfield{editor}{\bibinfo{person}{S.~Bengio}, \bibinfo{person}{H.~Wallach},
  \bibinfo{person}{H.~Larochelle}, \bibinfo{person}{K.~Grauman},
  \bibinfo{person}{N.~Cesa-Bianchi}, {and} \bibinfo{person}{R.~Garnett}}
  (Eds.), Vol.~\bibinfo{volume}{31}. \bibinfo{publisher}{Curran Associates,
  Inc.}
\newblock
\urldef\tempurl%
\url{https://proceedings.neurips.cc/paper_files/paper/2018/file/45a766fa266ea2ebeb6680fa139d2a3d-Paper.pdf}
\showURL{%
\tempurl}


\bibitem[Malik and Becker(2020)]%
        {malik2020}
\bibfield{author}{\bibinfo{person}{Osman~Asif Malik} {and}
  \bibinfo{person}{Stephen Becker}.} \bibinfo{year}{2020}\natexlab{}.
\newblock \showarticletitle{Guarantees for the {K}ronecker fast
  {J}ohnson--{L}indenstrauss transform using a coherence and sampling
  argument}.
\newblock \bibinfo{journal}{\emph{Linear Algebra Appl.}}  \bibinfo{volume}{602}
  (\bibinfo{year}{2020}), \bibinfo{pages}{120--137}.
\newblock
\showISSN{0024-3795}
\urldef\tempurl%
\url{https://doi.org/10.1016/j.laa.2020.05.004}
\showDOI{\tempurl}


\bibitem[Minster et~al\mbox{.}(2020)]%
        {minster}
\bibfield{author}{\bibinfo{person}{Rachel Minster}, \bibinfo{person}{Arvind~K.
  Saibaba}, {and} \bibinfo{person}{Misha~E. Kilmer}.}
  \bibinfo{year}{2020}\natexlab{}.
\newblock \showarticletitle{Randomized algorithms for low-rank tensor
  decompositions in the {T}ucker format}.
\newblock \bibinfo{journal}{\emph{SIAM Journal on Mathematics of Data Science}}
  \bibinfo{volume}{2}, \bibinfo{number}{1} (\bibinfo{year}{2020}),
  \bibinfo{pages}{189--215}.
\newblock
\urldef\tempurl%
\url{https://doi.org/10.1137/19M1261043}
\showDOI{\tempurl}


\bibitem[Rakhshan and Rabusseau(2020)]%
        {trp}
\bibfield{author}{\bibinfo{person}{Beheshteh Rakhshan} {and}
  \bibinfo{person}{Guillaume Rabusseau}.} \bibinfo{year}{2020}\natexlab{}.
\newblock \showarticletitle{Tensorized random projections}. In
  \bibinfo{booktitle}{\emph{Proceedings of the Twenty Third International
  Conference on Artificial Intelligence and Statistics}}
  \emph{(\bibinfo{series}{Proceedings of Machine Learning Research},
  Vol.~\bibinfo{volume}{108})}, \bibfield{editor}{\bibinfo{person}{Silvia
  Chiappa} {and} \bibinfo{person}{Roberto Calandra}} (Eds.).
  \bibinfo{publisher}{PMLR}, \bibinfo{pages}{3306--3316}.
\newblock


\bibitem[Tsourakakis({[n.\,d.]})]%
        {rand3}
\bibfield{author}{\bibinfo{person}{Charalampos~E. Tsourakakis}.}
  \bibinfo{year}{[n.\,d.]}\natexlab{}.
\newblock \bibinfo{booktitle}{\emph{{MACH}: Fast randomized tensor
  decompositions}}.
\newblock \bibinfo{pages}{689--700}.
\newblock
\urldef\tempurl%
\url{https://doi.org/10.1137/1.9781611972801.60}
\showDOI{\tempurl}
\showeprint{https://epubs.siam.org/doi/pdf/10.1137/1.9781611972801.60}


\bibitem[Wang et~al\mbox{.}(2011)]%
        {wang_data}
\bibfield{author}{\bibinfo{person}{Yalin Wang}, \bibinfo{person}{Yang Song},
  \bibinfo{person}{Priya Rajagopalan}, \bibinfo{person}{Tuo An},
  \bibinfo{person}{Krystal Liu}, \bibinfo{person}{Yi-Yu Chou},
  \bibinfo{person}{Boris Gutman}, \bibinfo{person}{Arthur~W. Toga}, {and}
  \bibinfo{person}{Paul~M. Thompson}.} \bibinfo{year}{2011}\natexlab{}.
\newblock \showarticletitle{Surface-based {TBM} boosts power to detect disease
  effects on the brain: An {N}=804 {ADNI} study}.
\newblock \bibinfo{journal}{\emph{NeuroImage}} \bibinfo{volume}{56},
  \bibinfo{number}{4} (\bibinfo{year}{2011}), \bibinfo{pages}{1993--2010}.
\newblock
\showISSN{1053-8119}
\urldef\tempurl%
\url{https://doi.org/10.1016/j.neuroimage.2011.03.040}
\showDOI{\tempurl}


\bibitem[Zhou et~al\mbox{.}(2013)]%
        {neuro_tensor}
\bibfield{author}{\bibinfo{person}{Hua Zhou}, \bibinfo{person}{Lexin Li}, {and}
  \bibinfo{person}{Hongtu Zhu}.} \bibinfo{year}{2013}\natexlab{}.
\newblock \showarticletitle{Tensor regression with applications in neuroimaging
  data analysis}.
\newblock \bibinfo{journal}{\emph{J. Amer. Statist. Assoc.}}
  \bibinfo{volume}{108}, \bibinfo{number}{502} (\bibinfo{year}{2013}),
  \bibinfo{pages}{540--552}.
\newblock
\urldef\tempurl%
\url{https://doi.org/10.1080/01621459.2013.776499}
\showDOI{\tempurl}
\showeprint{https://doi.org/10.1080/01621459.2013.776499}


\end{thebibliography}

\clearpage

\appendix

\section{Additional Background Results}


\begin{lemma} \label{lem2}
If $x,y\in\R^n$ and $\bA\in\R^{m\times n}$ is an $\varepsilon$-JL embedding of $\{x\pm y\}\subset\R^n$ into $\R^m$, then
\begin{align*}
    \big\lvert\lrangle{\bA x, \bA y}-\lrangle{x,y}\big\rvert &\leq \frac{\varepsilon}{2}(\lrVert{x}_2^2 + \lrVert{y}_2^2)\\
    &\leq \varepsilon\max\{\lrVert x_2^2,\lrVert y_2^2\}.
\end{align*}
\end{lemma}

\begin{proof}
The claim follows by routine manipulation. Observe that
\begin{align*}
    \big\lvert\lrangle{\bA x, \bA y}&-\lrangle{x,y}\big\rvert\\
    &= \Big\lvert \frac{1}{4}\big(\lrVert{\bA x + \bA y}_2^2 - \lrVert{\bA x - \bA y}_2^2\big)\\
    &\qquad\qquad - \frac{1}{4}\big(\lrVert{x+y}_2^2-\lrVert{x-y}_2^2\big)\Big\rvert \\
    &= \frac{1}{4} \Big\lvert \big(\lrVert{\bA x + \bA y}_2^2 - \lrVert{x+y}_2^2\big)\\
    &\qquad\qquad - \big(\lrVert{\bA x - \bA y}_2^2 - \lrVert{x-y}_2^2\big) \Big\rvert \\
    &\leq \frac{1}{4}\Big[ \big\lvert \lrVert{\bA x + \bA y}_2^2 - \lrVert{x+y}_2^2 \big\rvert\\
    &\qquad\qquad + \big\lvert \lrVert{\bA x - \bA y}_2^2 - \lrVert{x-y}_2^2 \big\rvert  \Big] \\
    &\leq \frac{1}{4}\big(\varepsilon\lrVert{x+y}_2^2 + \varepsilon\lrVert{x-y}_2^2\big)\\
    &= \frac{\varepsilon}{2}\big(\lrVert{x}_2^2+\lrVert{y}_2^2\big)\\
    &\leq \varepsilon\max\{\lrVert{x}_2^2,\lrVert{y}_2^2\},
\end{align*}
which proves the claim. Above, the first equality holds by the polarization identity and the inequalities by the triangle inequality, the hypothesis that $\bA$ is an $\varepsilon$-JL embedding, and by the parallelogram law (equivalently, by basic properties of inner products).
\end{proof}

The following is a proof of Lemma 2.1 of the main text.

\begin{proof}
    The claim follows by routine manipulation and the properties of tensor matricization. Observe that
    \begin{align*}
        \bY^\prime\mat{j} &= \bB\bY\mat{j}\\
        &= \bB\sumR\lr\big(\mathop{\bigcirc}_{k=1}^q\Gamma_{k,r_k}\big)\mat{j}\\
        &= \bB\sumR\lr\Gamma_{j,r_j}\Big(\mathop{\bigotimes}_{\substack{k=q\\k\neq j}}^{1}\Gamma_{k,r_k}\Big)^\top\\
        &= \sumR\lr(\bB\Gamma_{j,r_j})\Big(\mathop{\bigotimes}_{\substack{k=q\\k\neq j}}^{1}\Gamma_{k,r_k}\Big)^\top.
    \end{align*}
    Thus,
    \begin{align*}
      \bY^\prime = \sumR\lr &\Big(\mathop{\bigcirc}_{k=1}^{j-1}\Gamma_{k,r_k}\Big)\circ(\bB\Gamma_{j,r_j})\circ\\
      &\qquad\Big(\mathop{\bigcirc}_{k=j+1}^{q}\Gamma_{k,r_k}\Big).  
    \end{align*}
    Multiplying the general term above by $\lrVert{\bB\Gamma_{j,r_j}}/\lrVert{\bB\Gamma_{j,r_j}}$ (which is possible under the hypothesis that $\min_{r\in[R_j]}\lrVert{\bB\Gamma_{j,r}}_2>0$) yields the first part of the claim.

    From the form of $\bY^\prime$ above, it is clear that $\mu_{\bY^\prime,k}=\mu_{\bY,k}$ for $k\neq j$. On the other hand,
    $$\mu_{\bY^\prime,j} = \max_{\substack{r,s\in[R_j]\\r\neq s}}\frac{\lrvert{\lrangle{\bB\Gamma_{j,r},\bB\Gamma_{j,s}}}}{\lrVert{\bB\Gamma_{j,r}}_2\lrVert{\bB\Gamma_{j,s}}_2}.$$

    For the final part of the claim, observe that
    \begin{align*}
        \lrVert{\bY^\prime} &= \lrVert{\bY\times_k\bB}^2 = \lrVert{(\bY\times_j\bB)\mat{j}}_\text{F}^2\\
        &= \Big\lVert\bB\bG_j\bL\mat{j}\Big(\mathop{\bigotimes}_{\substack{k=q\\k\neq j}}^1\bG_k\Big)^\top\Big\rVert_\text{F}^2\\
        &= \lrVertF{\bB\bG_j\bP_j}^2.
    \end{align*}
    Thus, by direct computation,
    \begin{align*}
        \lrVert{\bY^\prime}^2
        &= \lrVertF{\bB\bG_j\bP_j}^2\\
        &= \sum_{i=1}^{N_{-j}}\big\lVert\bB\bG_j\Psi_i\big\rVert_2^2\\
        &= \sum_{i=1}^{N_{-j}}\Big\lVert\bB\sum_{r=1}^{R_j}\Gamma_{j,r}\psi_{r,i}\Big\rVert_2^2\\
        &= \sum_{i=1}^{N_{-j}}\lrangle{\bB\sum_{r=1}^{R_j}\Gamma_{j,r}\psi_{r,i},\bB\sum_{s=1}^{R_j}\Gamma_{j,s}\psi_{s,i}}\\
        &= \sum_{r=1}^{R_j}\sum_{s=1}^{R_j}\Big(\sum_{i=1}^{N_{-j}}\psi_{r,i}\psi_{s,i}\Big)\lrangle{\bB\Gamma_{j,r},\bB\Gamma_{j,s}}\\
        &= \sum_{r=1}^{R_j}\sum_{s=1}^{R_j}(\bP\bP^\top)_{r,s}\lrangle{\bB\Gamma_{j,r},\bB\Gamma_{j,s}}.
    \end{align*}
    Above, the second equality holds since, for any arbitrary $\bA\in\R^{n\times m}$, $\lrVertF{\bA}^2=\sum_{i=1}^m\lrVert{A_i}_2^2$. The third and final equalities simply use the definition of matrix multiplication. This completes the proof.
\end{proof}

\begin{figure*}[hb!]
    \centering
    \includegraphics[width=0.75\textwidth]{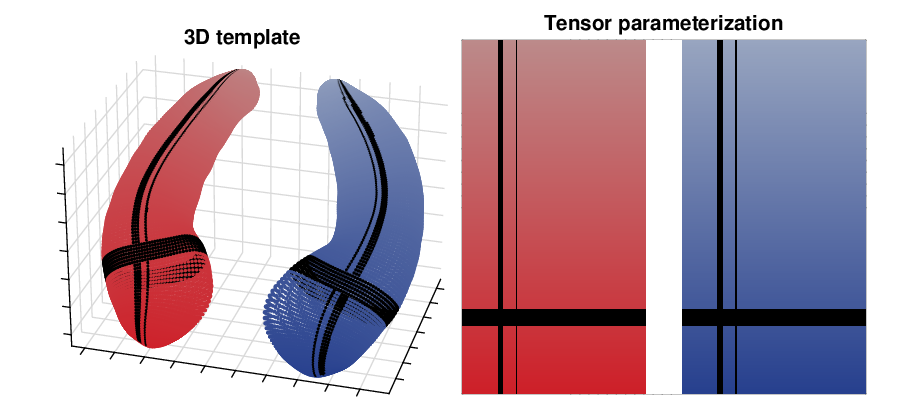}
    \caption{Visualization of the 3D template and $150\times100\times2$ parameterization used in the hippocampus analysis. The color gradients and solid black lines indicate the correspondence between the representations.} \label{fig:param}
    \Description{A two-panel figure. One panel shows a 3D image of a hippocampus and the other shows two rectangles. Red and blue gradient maps (respectively on the left and right hippocampus), along with black lines of different widths, illustrate the correspondence between the hippocampus surface and the tensor parameterization used in the empirical analysis of the hippocampus dataset.}
\end{figure*}

\begin{table*}[htb!]
{\centering
\caption{Mean (standard deviation) time (ms/iteration) spent on each part of the HOOI, HOOI-RE*, HOOI-RE, and HOSVD algorithms in both numerical studies, calculated over 100 simulations.} \label{tab:parts}
\setlength{\tabcolsep}{1.5pt}
\scalebox{0.85}{
\begin{tabular}{llrrrrrrrrr}
\hline
Method  & DR  & \multicolumn{3}{l}{$R = 5$}                & \multicolumn{3}{l}{$R = 15$}               & \multicolumn{3}{l}{$R = 30$}                 \\ \cline{3-11} 
        &     & $\bA$         & $\bG$        & $\bL$       & $\bA$         & $\bG$        & $\bL$       & $\bA$         & $\bG$         & $\bL$        \\ \midrule
\multicolumn{11}{c}{Face}                                                                                                                              \\ \hline
HOOI     & & 0.0 (0.0)     & 18.5 (1.3)   & 3.8 (0.6)   & 0.0 (0.0)     & 23.4 (2.2)   & 5.2 (0.8)   & 0.0 (0.0)     & 33.6 (3.4)    & 8.8 (0.9)    \\
HOOI-RE* & 0.8 & 21.9 (3.6)    & 14.8 (0.8)   & 4.2 (0.1)   & 22.2 (2.9)    & 14.4 (0.4)   & 4.7 (0.1)   & 23.7 (2.3)    & 19.2 (2.2)    & 9.9 (2.0)    \\
        & 0.6 & 14.0 (0.1)    & 8.1 (0.5)    & 3.5 (0.5)   & 14.8 (0.3)    & 8.7 (0.2)    & 4.8 (0.1)   & 15.4 (0.3)    & 10.3 (0.1)    & 8.7 (0.1)    \\
        & 0.4 & 9.2 (0.3)     & 4.9 (0.2)    & 3.6 (0.5)   & 9.2 (0.1)     & 5.1 (0.0)    & 4.6 (0.0)   & 10.0 (0.4)    & 6.6 (0.8)     & 10.3 (2.1)   \\
        & 0.2 & 5.2 (0.2)     & 2.7 (0.1)    & 3.0 (0.0)   & 5.0 (0.1)     & 2.6 (0.2)    & 4.7 (0.5)   & 6.0 (0.1)     & 4.0 (0.0)     & 11.4 (0.3)   \\
HOOI-RE  & 0.8 & 23.2 (3.4)    & 14.4 (0.6)   & 8.6 (0.2)   & 23.1 (2.8)    & 14.6 (0.2)   & 8.8 (0.1)   & 24.9 (3.1)    & 17.5 (0.3)    & 10.7 (0.1)   \\
        & 0.6 & 14.9 (0.3)    & 7.9 (0.3)    & 4.8 (0.1)   & 15.1 (0.3)    & 8.2 (0.1)    & 5.2 (0.1)   & 16.2 (0.3)    & 11.1 (1.1)    & 6.7 (0.3)    \\
        & 0.4 & 9.1 (0.1)     & 4.6 (0.1)    & 2.6 (0.0)   & 9.7 (0.3)     & 5.1 (0.1)    & 3.0 (0.0)   & 9.9 (0.3)     & 5.9 (0.1)     & 3.5 (0.1)    \\
        & 0.2 & 5.0 (0.1)     & 2.6 (0.2)    & 1.7 (0.2)   & 5.4 (0.1)     & 2.9 (0.1)    & 1.8 (0.0)   & 6.6 (0.6)     & 4.4 (0.1)     & 2.8 (0.2)    \\
HOSVD   & & 0.0 (0.0) & 445.7 (43.4) & 3.4 (0.4) & 0.0 (0.0) & 442.4 (39.5) & 5.1 (0.7) & 0.0 (0.0) & 440.6 (45.2) & 8.7 (1.6)\\
\hline
\multicolumn{11}{c}{Hippocampus}                                                                                                                       \\ \hline
HOOI     & 1.0 & 0.0 (0.0)     & 565.0 (44.8) & 64.6 (10.1) & 0.0 (0.0)     & 920.2 (6.9)  & 121.3 (1.3) & 0.0 (0.0)     & 1692.5 (17.8) & 255.9 (4.8)  \\
HOOI-RE* & 0.8 & 1676.0 (16.9) & 764.6 (16.1) & 57.9 (6.3)  & 1724.9 (20.5) & 802.8 (18.3) & 125.2 (6.7) & 1727.3 (13.5) & 823.6 (5.6)   & 242.4 (1.5)  \\
 & 0.6 & 934.9 (12.2)  & 379.2 (8.0)  & 64.0 (6.8)  & 967.6 (17.7)  & 408.3 (12.8) & 126.6 (9.3) & 999.5 (11.9)  & 417.4 (6.3)   & 245.7 (2.4)  \\
 & 0.4 & 407.2 (6.4)   & 148.8 (4.4)  & 61.5 (3.8)  & 398.8 (5.6)   & 150.2 (5.5)  & 121.5 (5.9) & 420.0 (5.6)   & 168.4 (11.1)  & 246.9 (2.3)  \\
 & 0.2 & 74.8 (2.1)    & 22.8 (1.2)   & 70.0 (9.4)  & 59.9 (1.1)    & 20.4 (1.1)   & 122.5 (3.0) & 77.5 (2.5)    & 29.4 (2.6)    & 274.0 (14.4) \\
HOOI-RE  & 0.8 & 1581.2 (18.1) & 739.0 (4.3)  & 268.3 (4.5) & 1665.9 (19.0) & 786.0 (19.8) & 306.1 (7.7) & 1838.3 (34.9) & 865.0 (16.9)  & 410.0 (11.6) \\
        & 0.6 & 887.1 (9.6)   & 358.9 (3.4)  & 136.0 (1.9) & 918.8 (9.5)   & 402.8 (19.1) & 153.5 (2.2) & 1052.1 (17.4) & 433.1 (14.1)  & 216.7 (3.4)  \\
        & 0.4 & 428.0 (4.2)   & 160.7 (7.1)  & 52.8 (1.7)  & 395.2 (3.7)   & 148.0 (1.8)  & 57.5 (1.7)  & 441.5 (7.0)   & 182.1 (10.5)  & 89.1 (4.1)   \\
        & 0.2 & 60.5 (2.7)    & 20.6 (1.5)   & 7.8 (0.4)   & 63.3 (1.3)    & 23.1 (1.1)   & 9.5 (0.3)   & 80.2 (4.6)    & 33.2 (2.1)    & 17.3 (0.6)   \\
HOSVD   & 1.0 & 0.0 (0.0) & 42242.4 (1032.7) & 108.7 (14.2) & 0.0 (0.0) & 42131.2 (861.3) & 219.4 (21.0) & 0.0 (0.0) & 42484.0 (1171.8) & 454.8 (10.9)\\
\hline
\end{tabular}
}}
{\\\footnotesize \textbf{Algorithm parts:} $\bA$, $\bG$, and $\bL$ denote the application of random embeddings, updates for the factor matrices, and updates for the core tensor, respectively.}\\
{\footnotesize \textbf{HOSVD:} HOSVD is not an iterative algorithm and terminates after a single update to each of the factor matrices and the core matrix.}\\
{\footnotesize \textbf{Omissions:} Time spent forming the random matrices is not provided for the sake of space; mean time for the HOOI-RE* and HOOI-RE algorithms across all settings ranged from 0.2 to 0.4 (with standard deviations from 0.0 to 0.1) ms/iteration. Time spent preprocessing is also not included, but is accounted for by other tables and figures in the main text. Specifically, time spent ``mixing'' the data took on average 75.9 (standard deviation 8.0) ms/replication in the face analysis and 7868.3 (standard deviation 746.7) ms/replication in the hippocampus analysis.}
\end{table*}

\end{document}